%% file: main.tex
\documentclass[10pt,twocolumn,letterpaper]{article}

\usepackage[pagenumbers]{cvpr} 

\usepackage[utf8]{inputenc} 
\usepackage[T1]{fontenc}    
\usepackage{url}            
\usepackage{booktabs}       
\usepackage{amsfonts}       
\usepackage{nicefrac}       
\usepackage{microtype}      
\usepackage{xcolor}         

\usepackage{verbatim}
\usepackage{bm}
\usepackage{bbm}
\usepackage{multirow}
\usepackage{titletoc}
\usepackage{array}
\usepackage{pifont}
\usepackage{nccmath}
\usepackage{titletoc}
\usepackage{subcaption}
\usepackage[table]{xcolor}

\usepackage{physics}
\usepackage{amsmath}
\usepackage{tikz}
\usepackage{mathdots}
\usepackage{yhmath}
\usepackage{cancel}
\usepackage{color}
\usepackage{siunitx}
\usepackage{array}
\usepackage{multirow}
\usepackage{amssymb}
\usepackage{gensymb}
\usepackage{tabularx}
\usepackage{extarrows}
\usepackage{enumitem}
\usepackage{booktabs}
\usetikzlibrary{fadings}
\usetikzlibrary{patterns}
\usetikzlibrary{shadows.blur}
\usetikzlibrary{shapes}
\usepackage{wrapfig}

\usepackage[linesnumbered,ruled,vlined]{algorithm2e}
\usepackage{chngcntr}
\usepackage{tocloft}
\usepackage[titletoc,toc,title]{appendix}

\usepackage{tcolorbox}
\tcbuselibrary{skins}

\input{preamble}
\definecolor{cvprblue}{rgb}{0.21,0.49,0.74}
\definecolor{purple}{rgb}{0.55,0.33,0.78}
\usepackage[pagebackref,breaklinks,colorlinks,anchorcolor=cvprblue, citecolor=cvprblue,linkcolor=cvprblue]{hyperref}

\usepackage[capitalize]{cleveref}
\crefname{table}{Table}{Tables}
\crefname{figure}{Figure}{Figures}
\crefname{algorithm}{Algorithm}{Algorithms}
\crefname{section}{Section}{Sections}
\crefname{appendix}{Appendix}{Appendices}
\crefname{equation}{Eq.}{Eqs.}
\crefname{assmp}{Assumption}{Assumptions}

\usepackage{amsthm}
\theoremstyle{plain}

\theoremstyle{definition}

\crefname{thm}{Theorem}{Theorems}
\crefname{rem}{Remark}{Remarks}
\crefname{prop}{Proposition}{Propositions}

\definecolor{soft_red}{RGB}{240, 110, 110}
\definecolor{soft_blue}{RGB}{100, 170, 220}
\definecolor{define_red}{RGB}{255, 20, 20}
\definecolor{define_green}{RGB}{41, 165, 31}

\newcommand{\pinput}[1]{\textcolor{blue!65!black}{#1}}
\newcommand{\poutput}[1]{\textcolor{orange!80!black}{#1}}

\def\paperID{35919} 
\def\confName{CVPR}
\def\confYear{2026}

\title{Training-Free Composed Video Retrieval via Visual \\ Representation-Guided Video-LLM Reasoning}

\author{
	Yang Liu\textsuperscript{1}\hspace{1.2em} Qianqian Xu\textsuperscript{2,3,}\thanks{Corresponding authors}\hspace{1.2em} Peisong Wen\textsuperscript{1}\hspace{1.2em} Siran Dai\textsuperscript{4,5}\hspace{1.2em} Qingming Huang\textsuperscript{1,2,*} \\
	{\textsuperscript{1}School of Computer Science and Technology, University of Chinese Academy of Sciences} \\
	{\textsuperscript{2}State Key Laboratory of AI Safety, Institute of Computing Technology, Chinese Academy of Sciences} \\
    {\textsuperscript{3}Beijing Academy of Artificial Intelligence} \\
    {\textsuperscript{4}Institute of Information Engineering, Chinese Academy of Sciences} \\
    {\textsuperscript{5}School of Cyber Security, University of Chinese Academy of Sciences} \\
	{\tt\small liuyang232@mails.ucas.ac.cn\hspace{2em} xuqianqian@ict.ac.cn\hspace{2em} wenpeisong@ucas.ac.cn } \\ 
    {\tt\small daisiran@iie.ac.cn\hspace{2em} qmhuang@ucas.ac.cn }
}

\begin{document}

\maketitle
\input{sec/0_abstract}    
\input{sec/1_introduction}
\input{sec/3_method}
\input{sec/4_experiments}
\input{sec/5_conclusion}

{
    \small
    \bibliographystyle{ieeenat_fullname}
    \bibliography{main}
}


\end{document}

%% file: sec/0_abstract.tex
\begin{abstract}
Recent advances in large vision-language models have expanded video retrieval from simple text-based search to more flexible scenarios, where users may specify the desired result through both visual examples and textual instructions. 
In the CVPR 2026 Reason-Aware Composed Video Retrieval Challenge, the system is required to retrieve a target video according to a reference video and a modification instruction. 
To address this task, we develop \textit{Visual Representation-Guided Video-LLM Reasoning for Training-Free Composed Video Retrieval}. 
Our framework first uses frozen DINOv3 models to obtain a compact set of visually relevant candidates, and then applies large vision-language models to evaluate whether each candidate satisfies the modification instruction. 
A final reasoning-based refinement is further performed on the top candidates to improve the first-ranked prediction. 
Without training, our system achieves \textbf{48.78} Recall@1 and \textbf{51.48} Recall@5 on the test set. 
Future work may further improve retrieval accuracy through stronger video-LLMs and detailed integration between visual representations and language reasoning.
\end{abstract}

%% file: sec/1_introduction.tex
\section{Introduction}
\label{sec:introduction}

Large vision-language models (VLM) have recently achieved substantial progress in multimodal perception, visual instruction following, and visual reasoning~\cite{flamingo,blip2,Gemini2,GPT4,qwen3vl}. 
These advances have broadened the scope of video retrieval from conventional text-based search to more interactive and compositional scenarios, where users specify a reference visual input together with a desired modification. 
Such capability is relevant to content-based video search~\cite{s2vs,liu2024not,jin2025causal} and video generation~\cite{Sora,MAGI1,liu2025bootstrapping}, where the desired result is often defined by a change relative to an existing video rather than by an isolated text query.

The CVPR 2026 Reason-Aware Composed Video Retrieval Challenge~\cite{covr} focuses on this setting. 
Given a reference video and a textual modification, the goal is to retrieve a target video that preserves the relevant content of the reference video while reflecting the requested change. 
This challenge highlights an important direction for video-language retrieval that can reason about the visual consequences of an edit, including changes in object states, action phases, scenes, camera views, and temporal cues.

In such a multimodal scenario, visual information is often highly redundant, and more precise visual understanding is needed to achieve reliable vision-language matching. 
Recently, the progress of self-supervised representation learning~\cite{MoCov1,MoCov3,MAE,ijepa,iBOT,tcore,dai2026exploring} has led to increasingly strong visual foundation models, such as DINO series~\cite{DINO,DINOv2,DINOv3}. 
These models have shown strong transferability across recognition~\cite{TimeSformer,VideoMAEv2,Kinetics,SSV2} and dense visual understanding tasks~\cite{SiamMAE,VideoMAC,co_settle,DAVIS17,JHMDB,VIP}. 
Their ability to produce discriminative representations for visual objects and scenes under diverse appearances makes them useful as a visual prior before expensive video-language reasoning.

Motivated by this observation, we propose \textit{Visual Representation-Guided Video-LLM Reasoning for Training-Free Composed Video Retrieval}. 
We first use frozen DINOv3 backbones to extract frame-level visual representations and construct a compact candidate pool with Chamfer similarity~\cite{visil}. 
We then apply Qwen3-VL models for modification-aware reranking, refinement, and reasoning trace generation. 
The whole pipeline is training-free on the dataset.

On the test set of the CoVR-R Challenge, our final submission achieves 48.78 Recall@1 and 51.48 Recall@5. 
The result suggests that combining visual foundation representations with video-LLM reasoning is a simple and effective direction for composed video retrieval. 
This system remains an initial attempt under limited time and computational resources, and future improvements may come from larger-scale VLMs and closer integration between intermediate visual representations and language-based reasoning.

%% file: sec/3_method.tex
\section{Method}
\label{sec:method}

Given a reference video and a textual modification instruction, CoVR-R requires retrieving a target video that reflects the requested change while preserving the relevant content of the reference video. 
We propose a training-free coarse-to-fine framework that combines frozen visual representation models with large VLMs.

For the $i$-th query, we denote the reference video as $\bm{V}^{r}_i$ and the modification instruction as $e_i$. 
Given a candidate gallery $\mathcal{G}=\{\bm{V}^{c}_j\}_{j=1}^{N}$, the goal is to rank candidates according to their compatibility with the modified target semantics implied by $(\bm{V}^{r}_i,e_i)$. 
Our method consists of three steps. 
First, DINOv3 ViT-Large (ViT-L)\footnote{ \scriptsize \url{ https://huggingface.co/facebook/dinov3-vitl16-pretrain-lvd1689m}} and DINOv3 ViT-7B\footnote{ \scriptsize \url{https://huggingface.co/facebook/dinov3-vit7b16-pretrain-lvd1689m}} are used to construct a compact visual candidate pool. 
Second, Qwen3-VL-8B-Instruct\footnote{ \scriptsize \url{https://huggingface.co/Qwen/Qwen3-VL-8B-Instruct}} reranks the candidates by jointly considering the reference video, the modification instruction, and each candidate video. 
Finally, Qwen3-VL-8B-Thinking\footnote{ \scriptsize \url{https://huggingface.co/Qwen/Qwen3-VL-8B-Thinking}} refines the top-ranked candidates and generates the reasoning trace required for test submission.

\subsection{Retrieval with Visual Representation Models}

Applying a large VLM to all reference-candidate pairs is computationally expensive. 
Since a valid target video usually remains related to the reference video in subjects, scenes, object appearance, or motion patterns, visual similarity provides a useful prior for candidate retrieval.
We therefore first use frozen visual representations to reduce the search space.

For a video $\bm{V}$, we uniformly sample $T$ frames:
\begin{equation}
    \bm{V}=\{\bm{x}_t\}_{t=1}^{T}, 
    \quad 
    \bm{x}_t \in \mathbb{R}^{H \times W \times C}.
\end{equation}
Each frame is resized, normalized, and fed into the DINOv3 image encoder. 
We use the CLS token as the frame representation:
\begin{equation}
    \bm{F}(\bm{V})=\{\bm{f}_t\}_{t=1}^{T}, 
    \quad 
    \bm{f}_t \in \mathbb{R}^{D}.
\end{equation}

For a reference video $\bm{V}^{r}_i$ and a candidate video $\bm{V}^{c}_j$, we compute their Chamfer similarity:
\begin{equation}
    S_{i,j}
    =
    \frac{1}{T_i}
    \sum_{t=1}^{T_i}
    \max_{1 \leq u \leq T_j}
    \operatorname{cos}
    \left(
        \bm{f}^{r}_{i,t},
        \bm{f}^{c}_{j,u}
    \right),
    \label{eq:chamfer_similarity}
\end{equation}
where $\operatorname{cos}(\cdot,\cdot)$ denotes cosine similarity. 
This score matches each reference frame to its closest candidate frame and averages the best matching scores, preserving frame-level evidence better than direct average pooling.

We compute Eq.~\eqref{eq:chamfer_similarity} with both DINOv3 backbones and obtain two ranked lists, $\mathcal{L}^{\mathrm{L}}_i$ and $\mathcal{L}^{\mathrm{7B}}_i$. 
The two lists are merged by preserving their ranking order and removing duplicate candidates. 
The source video is excluded, and the resulting pool $\mathcal{C}_i$ is passed to the subsequent process.

\subsection{Modification-Aware Compositional Reranking}

The DINOv3 retrieval stage mainly captures reference-candidate similarity, which provides a preliminary correlation. 
However, CoVR-R requires the candidate to not only preserve relevant reference content but also implement the modification instruction. 
We therefore use Qwen3-VL-8B-Instruct to evaluate the composed compatibility of each candidate.

For each query, we take the top $K_r$ candidates from $\mathcal{C}_i$, where $K_r=200$ in the implementation. 
For each candidate, the model receives $\bm{V}^{r}_i$, $e_i$, and $\bm{V}^{c}_j$, and outputs a matching score from $0$ to $100$ with a brief reason. 
The prompt is summarized as follows:

\begin{quote}
\small\itshape
Given a reference video and a modification instruction, rate how well the candidate video matches the expected result. 

Modification instruction: \pinput{\{modification\}} 

Rate the candidate video from 0 to 100 based on whether it preserves elements from the reference video that should be kept, correctly implements the requested modification, and is consistent in action, state, scene, background, and temporal content. 

Reply with ONLY a \poutput{number 0--100} on the first line, then a \poutput{brief reason} on the second line.
\end{quote}

The reranking score is denoted as
\begin{equation}
    r_{i,j}
    =
    \Phi_{\mathrm{inst}}
    \left(
        \bm{V}^{r}_i,
        e_i,
        \bm{V}^{c}_j
    \right),
    \label{eq:instruct_score}
\end{equation}
where $\Phi_{\mathrm{inst}}$ denotes Qwen3-VL-8B-Instruct. 
Candidates are sorted by $r_{i,j}$ in descending order. 
When scores are tied or unavailable, the visual retrieval order is used as a fallback.

\subsection{Top-rank Candidate Reasoning}

The Instruct reranker provides a modification-aware ordering, but visually similar top candidates may still differ in subtle action states or scene details. 
We therefore verify the top $K_t$ candidates with Qwen3-VL-8B-Thinking, where $K_t=10$. 
This step focuses on the most competitive candidates and does not expand the candidate pool.

The thinking prompt follows the similar scoring criteria but asks the model to examine the evidence more carefully before giving the final score:

\begin{quote}
\small\itshape
Given a reference video and a modification instruction, rate how well the candidate video matches the expected result. 

Modification instruction: \pinput{\{modification\}} 

Rate the candidate video from 0 to 100 based on whether it preserves elements from the reference video that should be kept, correctly implements the requested modification, and is consistent in action, state, scene, background, and temporal content. Think carefully step by step about each aspect before giving your final score. If this candidate were ranked \#1, explain clearly why it is the best match for the reference video and modification instruction. 

After your reasoning, reply with ONLY a \poutput{number 0--100} on a new line, then a \poutput{brief reason} on the next line.
\end{quote}

The refined score is written as
\begin{equation}
    \hat{r}_{i,j}
    =
    \Phi_{\mathrm{think}}
    \left(
        \bm{V}^{r}_i,
        e_i,
        \bm{V}^{c}_j
    \right),
    \quad j \leq K_t,
    \label{eq:thinking_score}
\end{equation}
where $\Phi_{\mathrm{think}}$ denotes Qwen3-VL-8B-Thinking. 
Only the top-$K_t$ candidates are reordered by $\hat{r}_{i,j}$, while the remaining candidates keep the order from the Instruct reranker.

Finally, we construct the official submission by selecting the top-50 candidates after verification. 
Duplicate candidates are removed, the source video is excluded, and all candidate identifiers are checked against the corresponding gallery. 
For the test split, the reasoning generated for the final top-1 candidate is used as the \texttt{reasoning\_trace}. If it is unavailable, we use the brief explanation from the Instruct reranker as a fallback.

%% file: sec/4_experiments.tex
\section{Experiments}
\label{sec:experiments}

\paragraph{Benchmark and metrics.}
We evaluate our method on the test split of the CoVR-R Challenge. 
The benchmark contains 2,634 validation queries and 301 test queries, with 4,333 videos in the candidate gallery from SSv2~\cite{SSV2} and WebVid~\cite{webvid}. 
Each query consists of a reference video and a textual modification instruction, and the system is required to retrieve the corresponding target video from the gallery. 
Following the official protocol, we report Recall@$K$ with $K \in \{1,5,10,50\}$. 
A prediction is counted as correct if the ground-truth target appears within the top-$K$ retrieved candidates.

\paragraph{Implementation.}
All models are used in a training-free manner without fine-tuning on the validation or test split. 
We use DINOv3 ViT-L and DINOv3 ViT-7B to extract frame-level visual representations and build the initial candidate pool. 
For each query, the top 200 candidates are reranked by Qwen3-VL-8B-Instruct according to the reference video, modification instruction, and candidate video. 
The top 10 candidates are then refined by Qwen3-VL-8B-Thinking, and the final submission keeps the top 50 candidates after removing duplicated or invalid entries.

\subsection{Main Results}

Table~\ref{tab:main_results} reports the test results of the three stages in our pipeline. 
This progressive comparison is designed to examine whether each component improves the final top prediction.

\begin{table}[t]
\centering
\caption{Main results on the test split of CoVR-R Challenge. The metrics are Recall@$K$ in percentage.}
\label{tab:main_results}
\resizebox{\linewidth}{!}{
\begin{tabular}{l|cccc}
\toprule
Method & R@1 & R@5 & R@10 & R@50 \\
\midrule
Step 1: DINOv3 Results Merge & 25.34 & 26.52 & 26.68 & 42.72 \\
Step 2: Qwen3-VL-Instruct Rerank & 35.27 & 38.80 & 38.98 & 51.64 \\
Step 3: Qwen3-VL-Thinking Refinement & \textbf{48.78} & \textbf{51.48} & \textbf{51.48} & \textbf{51.64} \\
\bottomrule
\end{tabular}
}
\end{table}

The DINOv3 retrieval stage achieves 25.34 Recall@1. 
This result shows that strong visual representations provide a useful starting point for candidate retrieval, since the target video often retains visual elements from the reference video. 
However, visual similarity alone cannot reliably determine the final answer, because the top candidate must also satisfy the modification instruction.

Adding Qwen3-VL-8B-Instruct reranking improves Recall@1 from 25.34 to 35.27. 
This gain comes from introducing an explicit comparison among the reference video, the modification instruction, and each candidate video. 
The result suggests that video-LLM reasoning helps correct cases where visually similar candidates fail to reflect the requested semantic change.

The final Qwen3-VL-8B-Thinking refinement further increases Recall@1 from 35.27 to 48.78. 
This stage focuses on the most competitive candidates and performs a more careful check of their visual and semantic evidence. 
The improvement indicates that many remaining errors after reranking are not due to missing candidates, but due to subtle ordering mistakes among plausible top candidates.

%% file: sec/5_conclusion.tex
\section{Conclusion and Discussion}
\label{sec:conclusion}

In this work, we present \textit{Visual Representation-Guided Video-LLM Reasoning for Training-Free Composed Video Retrieval} for the CVPR 2026 Reason-Aware Composed Video Retrieval Challenge. 
Our method first uses frozen DINOv3 representations to select visually relevant candidates from the large gallery, and then applies VLMs to judge whether these candidates satisfy the modification instruction. 
The final top-rank refinement further improves the ordering of the most plausible candidates. 

This work is still an initial attempt under limited time and computational resources. 
Further improvements may come from stronger VLMs, more dedicated scoring criteria for candidate evaluation, and a closer integration between visual foundation models and language reasoning.

%% file: main.bib
@String(CVPR= {IEEE Conf. Comput. Vis. Pattern Recog.})

@String(ICCV= {Int. Conf. Comput. Vis.})

@String(NIPS= {Adv. Neural Inform. Process. Syst.})

@String(TIP  = {IEEE Trans. Image Process.})

@String(ACMMM= {ACM Int. Conf. Multimedia})

@String(ICLR = {Int. Conf. Learn. Represent.})

@String(CVPR  = {CVPR})

@String(ICCV  = {ICCV})

@String(NIPS  = {NeurIPS})

@String(TIP   = {IEEE TIP})

@String(ACMMM = {ACM MM})

@String(ICLR  = {ICLR})

@article{covr,
  title={Covr-r: Reason-aware composed video retrieval},
  author={Thawakar, Omkar and Demidov, Dmitry and Potlapalli, Vaishnav and Bogireddy, Sai Prasanna Teja Reddy and Gajjala, Viswanatha Reddy and Lasheen, Alaa Mostafa and Anwer, Rao Muhammad and Khan, Fahad},
  journal={arXiv preprint arXiv:2603.20190},
  year={2026}
}

@inproceedings{s2vs,
  title={Self-supervised video similarity learning},
  author={Kordopatis-Zilos, Giorgos and Tolias, Giorgos and Tzelepis, Christos and Kompatsiaris, Ioannis and Patras, Ioannis and Papadopoulos, Symeon},
  booktitle={Proceedings of the IEEE/CVF conference on computer vision and pattern recognition},
  pages={4756--4766},
  year={2023}
}

@inproceedings{visil,
  title={Visil: Fine-grained spatio-temporal video similarity learning},
  author={Kordopatis-Zilos, Giorgos and Papadopoulos, Symeon and Patras, Ioannis and Kompatsiaris, Ioannis},
  booktitle=ICCV,
  pages={6351--6360},
  year={2019}
}

@article{jin2025causal,
  title={Causal Inference Hashing for Long-Tailed Image Retrieval},
  author={Jin, Lu and Lu, Zhengyun and Li, Zechao and Pan, Yonghua and Dai, Longquan and Tang, Jinhui and Jain, Ramesh},
  journal=TIP,
  year={2025},
  publisher={IEEE}
}

@inproceedings{DINO,
  title={Emerging properties in self-supervised vision transformers},
  author={Caron, Mathilde and Touvron, Hugo and Misra, Ishan and J{\'e}gou, Herv{\'e} and Mairal, Julien and Bojanowski, Piotr and Joulin, Armand},
  booktitle=ICCV,
  pages={9650--9660},
  year={2021}
}

@inproceedings{MAE,
  title={Masked autoencoders are scalable vision learners},
  author={He, Kaiming and Chen, Xinlei and Xie, Saining and Li, Yanghao and Doll{\'a}r, Piotr and Girshick, Ross},
  booktitle=CVPR,
  pages={16000--16009},
  year={2022}
}

@inproceedings{MoCov1,
  title={Momentum contrast for unsupervised visual representation learning},
  author={He, Kaiming and Fan, Haoqi and Wu, Yuxin and Xie, Saining and Girshick, Ross},
  booktitle=CVPR,
  pages={9729--9738},
  year={2020}
}

@inproceedings{MoCov3,
  title={An empirical study of training self-supervised vision transformers},
  author={Chen, Xinlei and Xie, Saining and He, Kaiming},
  booktitle=ICCV,
  pages={9640--9649},
  year={2021}
}

@inproceedings{ijepa,
  title={Self-supervised learning from images with a joint-embedding predictive architecture},
  author={Assran, Mahmoud and Duval, Quentin and Misra, Ishan and Bojanowski, Piotr and Vincent, Pascal and Rabbat, Michael and LeCun, Yann and Ballas, Nicolas},
  booktitle=CVPR,
  pages={15619--15629},
  year={2023}
}

@article{SiamMAE,
  title={Siamese masked autoencoders},
  author={Gupta, Agrim and Wu, Jiajun and Deng, Jia and Li, Fei-Fei},
  journal=NIPS,
  volume={36},
  pages={40676--40693},
  year={2023}
}

@inproceedings{VideoMAC,
  title={VideoMAC: Video Masked Autoencoders Meet ConvNets},
  author={Pei, Gensheng and Chen, Tao and Jiang, Xiruo and Liu, Huafeng and Sun, Zeren and Yao, Yazhou},
  booktitle=CVPR,
  pages={22733--22743},
  year={2024}
}

@inproceedings{VideoMAEv2,
  title={Videomae v2: Scaling video masked autoencoders with dual masking},
  author={Wang, Limin and Huang, Bingkun and Zhao, Zhiyu and Tong, Zhan and He, Yinan and Wang, Yi and Wang, Yali and Qiao, Yu},
  booktitle=CVPR,
  pages={14549--14560},
  year={2023}
}

@inproceedings{TimeSformer,
  title={Is space-time attention all you need for video understanding?},
  author={Bertasius, Gedas and Wang, Heng and Torresani, Lorenzo},
  booktitle=ICML,
  year={2021}
}

@article{Kinetics,
  title={The kinetics human action video dataset},
  author={Kay, Will and Carreira, Joao and Simonyan, Karen and Zhang, Brian and Hillier, Chloe and Vijayanarasimhan, Sudheendra and Viola, Fabio and Green, Tim and Back, Trevor and Natsev, Paul and others},
  journal={arXiv preprint arXiv:1705.06950},
  year={2017}
}

@inproceedings{SSV2,
  title={The" something something" video database for learning and evaluating visual common sense},
  author={Goyal, Raghav and Ebrahimi Kahou, Samira and Michalski, Vincent and Materzynska, Joanna and Westphal, Susanne and Kim, Heuna and Haenel, Valentin and Fruend, Ingo and Yianilos, Peter and Mueller-Freitag, Moritz and others},
  booktitle=ICCV,
  pages={5842--5850},
  year={2017}
}

@article{DAVIS17,
  title={The 2017 davis challenge on video object segmentation},
  author={Pont-Tuset, Jordi and Perazzi, Federico and Caelles, Sergi and Arbel{\'a}ez, Pablo and Sorkine-Hornung, Alex and Van Gool, Luc},
  journal={arXiv preprint arXiv:1704.00675},
  year={2017}
}

@inproceedings{JHMDB,
  title={Towards understanding action recognition},
  author={Jhuang, Hueihan and Gall, Juergen and Zuffi, Silvia and Schmid, Cordelia and Black, Michael J},
  booktitle={Proceedings of the IEEE international conference on computer vision},
  pages={3192--3199},
  year={2013}
}

@inproceedings{VIP,
  title={Adaptive temporal encoding network for video instance-level human parsing},
  author={Zhou, Qixian and Liang, Xiaodan and Gong, Ke and Lin, Liang},
  booktitle={Proceedings of the 26th ACM international conference on Multimedia},
  pages={1527--1535},
  year={2018}
}

@inproceedings{webvid,
  title={Frozen in time: A joint video and image encoder for end-to-end retrieval},
  author={Bain, Max and Nagrani, Arsha and Varol, G{\"u}l and Zisserman, Andrew},
  booktitle=ICCV,
  pages={1728--1738},
  year={2021}
}

@article{iBOT,
  title={ibot: Image bert pre-training with online tokenizer},
  author={Zhou, Jinghao and Wei, Chen and Wang, Huiyu and Shen, Wei and Xie, Cihang and Yuille, Alan and Kong, Tao},
  journal=ICLR,
  year={2022}
}

@article{DINOv2,
  title={Dinov2: Learning robust visual features without supervision},
  author={Oquab, Maxime and Darcet, Timoth{\'e}e and Moutakanni, Th{\'e}o and Vo, Huy and Szafraniec, Marc and Khalidov, Vasil and Fernandez, Pierre and Haziza, Daniel and Massa, Francisco and El-Nouby, Alaaeldin and others},
  journal={Transactions on Machine Learning Research},
  year={2024},
}

@article{DINOv3,
  title={Dinov3},
  author={Sim{\'e}oni, Oriane and Vo, Huy V and Seitzer, Maximilian and Baldassarre, Federico and Oquab, Maxime and Jose, Cijo and Khalidov, Vasil and Szafraniec, Marc and Yi, Seungeun and Ramamonjisoa, Micha{\"e}l and others},
  journal={arXiv preprint arXiv:2508.10104},
  year={2025}
}

@article{flamingo,
  title={Flamingo: a visual language model for few-shot learning},
  author={Alayrac, Jean-Baptiste and Donahue, Jeff and Luc, Pauline and Miech, Antoine and Barr, Iain and Hasson, Yana and Lenc, Karel and Mensch, Arthur and Millican, Katherine and Reynolds, Malcolm and others},
  journal=NIPS,
  volume={35},
  pages={23716--23736},
  year={2022}
}

@inproceedings{blip2,
  title={Blip-2: Bootstrapping language-image pre-training with frozen image encoders and large language models},
  author={Li, Junnan and Li, Dongxu and Savarese, Silvio and Hoi, Steven},
  booktitle=ICML,
  pages={19730--19742},
  year={2023},
  organization={PMLR}
}

@article{GPT4,
  title={Gpt-4 technical report},
  author={Achiam, Josh and Adler, Steven and Agarwal, Sandhini and Ahmad, Lama and Akkaya, Ilge and Aleman, Florencia Leoni and Almeida, Diogo and Altenschmidt, Janko and Altman, Sam and Anadkat, Shyamal and others},
  journal={arXiv preprint arXiv:2303.08774},
  year={2023}
}

@article{Gemini2,
  title={Gemini 2.5: Pushing the frontier with advanced reasoning, multimodality, long context, and next generation agentic capabilities},
  author={Comanici, Gheorghe and Bieber, Eric and Schaekermann, Mike and Pasupat, Ice and Sachdeva, Noveen and Dhillon, Inderjit and Blistein, Marcel and Ram, Ori and Zhang, Dan and Rosen, Evan and others},
  journal={arXiv preprint arXiv:2507.06261},
  year={2025}
}

@article{Sora,
  title={Sora: A review on background, technology, limitations, and opportunities of large vision models},
  author={Liu, Yixin and Zhang, Kai and Li, Yuan and Yan, Zhiling and Gao, Chujie and Chen, Ruoxi and Yuan, Zhengqing and Huang, Yue and Sun, Hanchi and Gao, Jianfeng and others},
  journal={arXiv preprint arXiv:2402.17177},
  year={2024}
}

@article{MAGI1,
  title={MAGI-1: Autoregressive Video Generation at Scale},
  author={Teng, Hansi and Jia, Hongyu and Sun, Lei and Li, Lingzhi and Li, Maolin and Tang, Mingqiu and Han, Shuai and Zhang, Tianning and Zhang, WQ and Luo, Weifeng and others},
  journal={arXiv preprint arXiv:2505.13211},
  year={2025}
}

@article{qwen3vl,
  title={Qwen3-vl technical report},
  author={Bai, Shuai and Cai, Yuxuan and Chen, Ruizhe and Chen, Keqin and Chen, Xionghui and Cheng, Zesen and Deng, Lianghao and Ding, Wei and Gao, Chang and Ge, Chunjiang and others},
  journal={arXiv preprint arXiv:2511.21631},
  year={2025}
}

@inproceedings{liu2024not,
  title={Not All Pairs are Equal: Hierarchical Learning for Average-Precision-Oriented Video Retrieval},
  author={Liu, Yang and Xu, Qianqian and Wen, Peisong and Dai, Siran and Huang, Qingming},
  booktitle=ACMMM,
  pages={3828--3837},
  year={2024}
}

@inproceedings{tcore,
  title={When the future becomes the past: Taming temporal correspondence for self-supervised video representation learning},
  author={Liu, Yang and Xu, Qianqian and Wen, Peisong and Dai, Siran and Huang, Qingming},
  booktitle=CVPR,
  pages={24033--24044},
  year={2025}
}

@inproceedings{co_settle,
  title={From Static to Dynamic: Exploring Self-supervised Image-to-Video Representation Transfer Learning},
  author={Liu, Yang and Xu, Qianqian and Wen, Peisong and Dai, Siran and Zhao, Xilin and Huang, Qingming},
  booktitle=CVPR,
  pages={31250--31261},
  year={2026}
}

@article{liu2025bootstrapping,
  title={Bootstrapping Physics-Grounded Video Generation through VLM-Guided Iterative Self-Refinement},
  author={Liu, Yang and Zhao, Xilin and Wen, Peisong and Dai, Siran and Huang, Qingming},
  journal={arXiv preprint arXiv:2511.20280},
  year={2025}
}

@article{dai2026exploring,
  title={Exploring structural degradation in dense representations for self-supervised learning},
  author={Dai, Siran and Xu, Qianqian and Wen, Peisong and Liu, Yang and Huang, Qingming},
  journal=NIPS,
  volume={38},
  pages={16715--16764},
  year={2026}
}
